\documentclass[lettersize,twoside,journal]{IEEEtran}
\usepackage[utf8]{inputenc}
\usepackage[T1]{fontenc}
\usepackage{url}
\usepackage{graphicx}
\usepackage{amsmath}
\usepackage{amssymb}
\usepackage{amsthm}

  \theoremstyle{definition}
  
  \theoremstyle{remark}

\usepackage{booktabs}
\usepackage{multirow}
\usepackage{array}
\usepackage{tabularx}

\usepackage{makecell}
\usepackage{xltabular}
\usepackage{algorithm}
\usepackage{algorithmic}
\usepackage{stfloats}
\usepackage{textcomp}
\usepackage{cite}
\usepackage{listings}
\usepackage{CJKutf8}
\usepackage[most]{tcolorbox}
\definecolor{lightgray}{gray}{0.95}
\newcolumntype{C}[1]{>{\centering\arraybackslash}m{#1}}
\usepackage{hyperref}
\usepackage{balance}

\title{BiGraph-Diffuse: A Bidirectional Diffusion Language Model with Graph-Structured Retrieval For Mental Health Counseling}

\author{Yuxiang Cheng$^1$,~Quanwei Tang$^2$,~Lvhui Lu$^2$,~Dong Zhang$^2$, \textit{Member, IEEE},~Shoushan Li$^2$,~and~Erik Cambria$^3$, \textit{Fellow, IEEE} 
\thanks{$^1$Dongwu Business School, Soochow University, Suzhou, China. $^2$School of Computer Science and Technology, Soochow University, Suzhou, China. $^3$College of Computing and Data Science, Nanyang Technological University, Singapore}
\thanks{Manuscript received XXX; revised XXX.}}

\begin{document}

\maketitle
\thispagestyle{headings}

\begin{abstract}
Mental health disorders affect hundreds of millions of people around the world, yet access to professional counseling remains severely limited. AI-powered dialogue systems offer a scalable alternative, but existing models face two fundamental challenges. First, they lack the bidirectional understanding needed to capture the layered nature of emotional expression, particularly in cases of progressive disclosure, where clients often present symptoms at the surface-level while concealing deeper trauma. Autoregressive (AR) models process information sequentially and cannot revise early interpretations when new evidence emerges later in the conversation. Second, they fail to effectively incorporate the relational knowledge that underlies clinical reasoning.
In this paper, we propose \textbf{BiGraph-Diffuse}, the first large-scale diffusion language model tailored for the counseling domain. We further introduce \textbf{BiGraph-RAG}, a relation-free graph-structured retrieval strategy that relies only on lightweight entity extraction and semantic linking. This design preserves inferential pathways from observable symptoms to potential underlying causes, while incurring zero LLM token cost during indexing.
Importantly, these two modules are not merely combined but mutually reinforcing. The diffusion model provides a holistic bidirectional context, enabling the system to defer premature judgments during progressive disclosure. Meanwhile, graph-based retrieval captures the structured interconnections of clinical knowledge. Extensive experiments demonstrate the effectiveness of BiGraph-Diffuse, and we further provide a solid theoretical analysis to support its design.
\end{abstract}


\section{Introduction}
\IEEEPARstart{M}{ental} health disorders affect one in eight individuals worldwide, yet access to professional counseling remains severely limited due to resource constraints and social stigma. AI-powered dialogue systems offer a promising solution, but counseling poses challenges beyond general dialogue generation. Effective counseling requires not only empathetic responses but also a deep understanding of client narratives and the ability to leverage structured clinical knowledge~\cite{chung2023challenges}.

Existing AI-based counseling systems~\cite{yang2025cami, lu2026mctsrzero} suffer from a fundamental limitation: they treat psychological conversations as conventional question-answering tasks, assuming that users disclose their problems directly and linearly. In reality, psychological disclosure is often a process of \textbf{progressive disclosure} or \textbf{defensive narration}. Clients rarely reveal deep trauma at the outset. Instead, they present surface-level symptoms, such as insomnia or academic stress, that mask the underlying distress.

\begin{figure*}[!t]
    \centering
    \includegraphics[width=\textwidth, trim=0 17cm 0 1cm, clip]{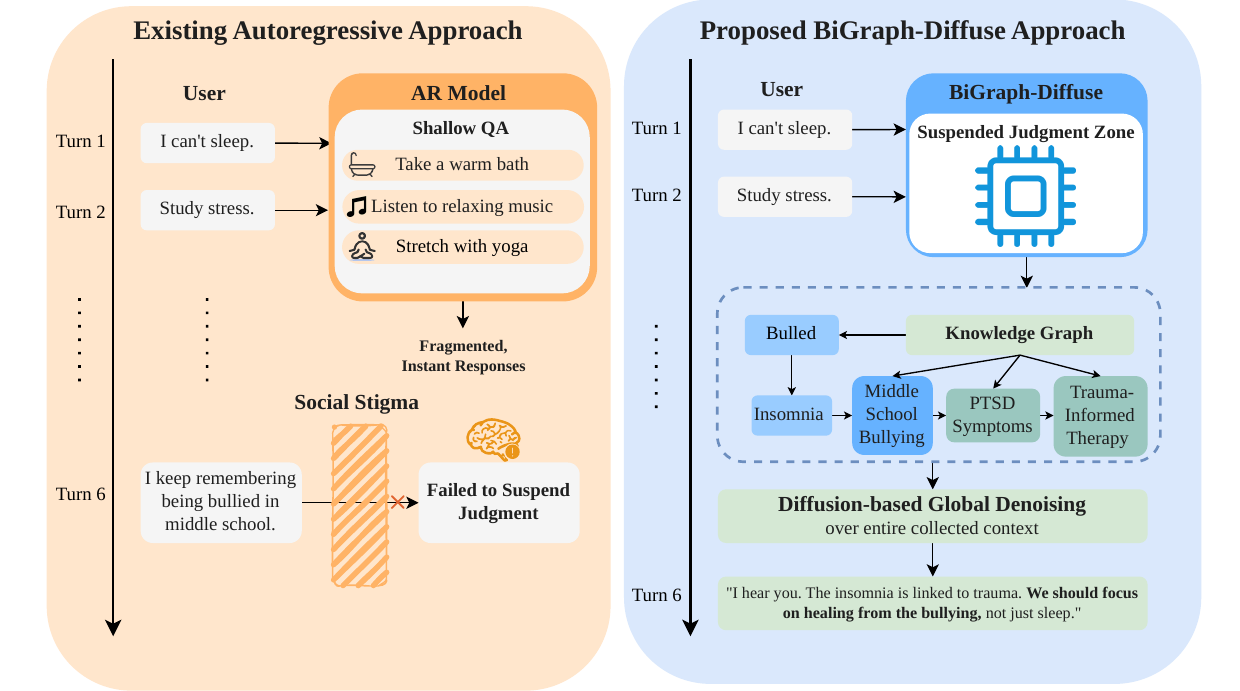}
    \caption{Motivation of BiGraph-Diffuse. Left: Autoregressive models fail to suspend judgment on early defensive disclosures (e.g., insomnia) and cannot revise shallow advice when deep trauma (e.g., bullying) is later revealed. Right: Traditional RAG retrieves isolated chunks and breaks the inferential chain between surface symptoms and underlying causes.}
    \label{fig:motivation}
\end{figure*}

This characteristic poses a major challenge for dominant autoregressive (AR) models~\cite{nguyen2025large, badawi2026trust}. They normally generate responses through left-to-right processing and cannot revise earlier interpretations when later evidence emerges. For example, as illustrated in the left panel of Figure~\ref{fig:motivation}, an AR model can initially respond to “I cannot sleep” with generic advice such as "take a warm bath" or "listen to relaxing music", only to discover several turns later (No. 6) that the symptom is due to school bullying. By then, its earlier shallow response cannot be corrected, which could make patients feel neglected, potentially causing harm. This architectural mismatch makes AR models ill-suited for the non-linear emotional narratives common in counseling.

To address this limitation, we introduce diffusion language models (DLMs)~\cite{nie2025large}. Unlike AR models, DLMs generate responses through iterative denoising over the full dialogue context, including clues revealed later in the conversation. This enables holistic understanding before committing to specific responses, mirroring a counselor’s ability to suspend judgment until sufficient context is available.
However, \textbf{contextual understanding alone is insufficient}. Once a deeper clue emerges, the model must connect it to earlier symptoms and retrieve relevant clinical knowledge. Traditional retrieval-augmented generation (RAG) methods~\cite{zhuang2025linearrg} often retrieve isolated text chunks, breaking the inferential links between surface symptoms, underlying causes, and intervention strategies. 

To handle this, we design \textbf{BiGraph-RAG}, a graph-structured retrieval \underline{strategy} that constructs lightweight semantic links among clinical entities and preserves these inferential pathways. For example, in the right panel of Figure~\ref{fig:motivation}, upon receiving the query “school bullying,” \textbf{BiGraph-RAG} would not retrieve isolated text chunks such as “how to deal with school bullying,” preserving the inferential chain that links “school bullying” to the earlier “insomnia” symptom and to PTSD.

Based on these insights, we propose \textbf{BiGraph-Diffuse}, a counseling dialogue generation \underline{framework} that integrates bidirectional diffusion modeling with graph-structured retrieval. These two components are \textbf{tightly coupled} rather than simply combined. The diffusion model provides global contextual reasoning that defers premature judgments, while BiGraph-RAG injects structured relational signals to guide denoising toward clinically grounded interpretations. Their interaction forms a mutually reinforcing loop, where holistic understanding improves graph evidence integration, and graph retrieval sharpens the diffusion process through coherent inferential pathways. To the best of our knowledge, this is the first work to adapt a large-scale diffusion language model to the mental health domain while integrating graph retrieval for clinical reasoning. To further validate generalizability across languages and cultural contexts, we augment training with the Chinese counseling dataset CPsyCounD~\cite{zhang2024cpsycoun} and evaluate on its corresponding test set CPsyCounE, in addition to the original OpenR1-Psy~\cite{hu2025beyond} benchmark.
Our main contributions are as follows:
$\bullet$ We propose \textbf{BiGraph-Diffuse}, the first framework that adapts a large-scale diffusion language model to mental health counseling. The bidirectional architecture enables holistic context understanding, allowing the model to suspend premature judgment during progressive disclosure---a capability that autoregressive models fundamentally lack and that cannot be recovered through prompt engineering alone.

$\bullet$ We design \textbf{BiGraph-RAG}, a relation-free graph-structured retrieval strategy that constructs a hierarchical entity--sentence--passage graph using only lightweight NER and semantic linking, incurring zero LLM token cost during indexing while preserving clinical inferential pathways from surface symptoms to underlying causes.

$\bullet$ We provide a rigorous theoretical analysis (see Appendix) establishing that the MDM generator combined with BiGraph-RAG achieves a strictly tighter variational lower bound than AR models with dense retrieval, with the advantage maximized on progressive-disclosure dialogues.

$\bullet$ Through extensive experiments on English (OpenR1-Psy) and Chinese (CPsyCounE) benchmarks---including automatic evaluation, human assessment by licensed clinicians, and cross-layer probing---we demonstrate that BiGraph-Diffuse significantly outperforms strong AR baselines. Controlled ablation studies further confirm that the bidirectional architecture, not prompt engineering, drives the performance gains, and that BiGraph-RAG consistently outperforms both dense and linear RAG alternatives under identical conditions.

\section{Related work}

\textbf{Traditional mental health dialogue systems.}
AI-powered mental health support has evolved from single-turn empathy generation~\cite{liu2023chatcounselor, sun2021psyqa} to multi-turn emotional support~\cite{qiu2024smile, chen2023soulchat} and therapeutic reasoning integration~\cite{zhang2024escot, yang2024mentallama, hu2025beyond}. Comprehensive surveys~\cite{wulanguage, ji2023rethinking} cover the landscape of LLM-based mental healthcare, while recent systems~\cite{yang2025cami, lu2026mctsrzero, zhu2026psiarena} and multimodal fusion approaches~\cite{zhu2026uncertainty} have advanced clinical interaction quality. However, existing systems share two limitations: (1) autoregressive processing cannot capture layered emotional expression under progressive disclosure; (2) independent passage retrieval misses relational clinical knowledge. These motivate our diffusion-based architecture and graph-structured retrieval. Recent datasets~\cite{qi-etal-2025-kokorochat, qiu-lan-2025-psydial, zhang2024cpsycoun, Yin_Li_Zhang_Wang_Shao_Li_Chen_Jiang_2025} further support this domain.

\textbf{Diffusion language models for mental health.}
Masked diffusion models (MDMs)~\cite{shi2024simplified, sahoo2024simple} generate text by iterative unmasking with bidirectional context. LLaDA~\cite{nie2025large} scaled MDMs to 8B parameters, matching strong AR models while overcoming the reversal curse~\cite{berglund2023reversal}. Domain-specific pretraining has proven effective for mental health~\cite{ji2023domain}, yet adapting diffusion models to this domain remains unexplored. To our knowledge, this work is the first to adapt a large-scale diffusion language model to mental health counseling. The bidirectional architecture mirrors clinical reasoning---counselors holistically consider how later disclosures reframe earlier statements. As we show empirically (Section~\ref{sec:suspend}), this advantage cannot be replicated by prompting AR models to ``suspend judgment,'' confirming bidirectionality as a structural necessity.

\textbf{Retrieval-augmented generation for mental health knowledge.}
Standard RAG fails to capture relational clinical knowledge. GraphRAG systems address this: Microsoft GraphRAG~\cite{edge2024from}, LightRAG~\cite{guo2024lightrag}, HippoRAG~\cite{gutierrez2024hipporag}, PruneRAG~\cite{jiao2026prunerag}, and LinearRAG~\cite{zhuang2025linearrg}. However, most rely on explicit relation extraction---unstable for clinical text and expensive via LLM calls. Inspired by LinearRAG's relation-free design, we propose BiGraph-RAG, using only entity extraction and semantic linking with zero LLM token cost. Its two-stage retrieval---local semantic bridging followed by global importance aggregation---mimics a counselor's associative memory. This is the first integration of graph-structured RAG with a diffusion language model for mental health counseling.

\textbf{Evaluation of mental health dialogue systems.}
Reliable evaluation remains challenging, as lexical metrics correlate poorly with therapeutic quality~\cite{liu2023chatcounselor}. LLM-based judges~\cite{guo2024large} now assess empathy, safety, and clinical appropriateness, validated via human correlation~\cite{badawi2026trust}, with benchmarks like PsyArena~\cite{zhu2026psiarena} providing standardized protocols. Our work extends this by introducing progressive-disclosure-specific evaluation (integration, revision, sensitivity, temporal consistency), validating against licensed clinical psychologists ($r \ge 0.674$), cross-validating with a secondary judge, and reporting scores with and without reasoning traces.

\section{Methodology}

\label{sec:methodology}

\begin{figure*}[!t]
    \centering
    \includegraphics[width=\textwidth, trim=0 18cm 0 1cm, clip]{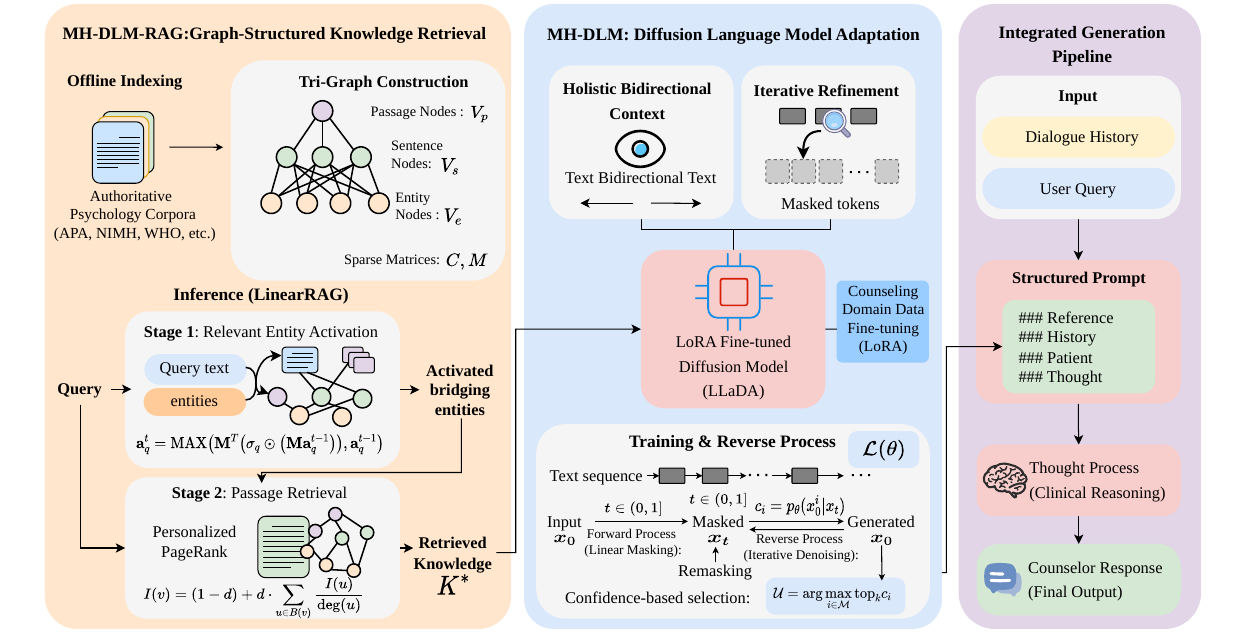}
    \caption{Overview of the BiGraph-Diffuse framework. The system consists of three stages: (1) Offline indexing constructs a heterogeneous Tri-Graph from authoritative psychology corpora; (2) During inference, BiGraph-RAG performs two-stage propagation over the graph to retrieve relevant knowledge; (3) The fine-tuned BiGraph-Diffuse model generates a thought process and final response conditioned on both the retrieved knowledge and dialogue history.}
    \label{fig:architecture}
\end{figure*}

Our proposed framework consists of three stages, depicted in Figure~\ref{fig:architecture}. \textbf{Stage 1 (Offline Indexing):} A heterogeneous Tri-Graph is constructed from 500 authoritative psychology passages (Section~\ref{sec:bigraph_rag}; also detailed in Appendix), linking entity, sentence, and passage nodes without explicit relation extraction. \textbf{Stage 2 (Online Retrieval):} Given a dialogue context, BiGraph-RAG performs two-stage propagation---semantic entity activation followed by personalized PageRank for passage retrieval---to obtain the top-$k$ clinically relevant passages. \textbf{Stage 3 (Generation):} The fine-tuned BiGraph-Diffuse model generates a structured output (clinical reasoning trace followed by counselor response) conditioned on the full dialogue history and retrieved knowledge. The diffusion architecture's bidirectional conditioning ensures that retrieval evidence can influence generation at any point in the denoising trajectory. We detail each component below.

\subsection{BiGraph-Diffuse-Base: adapting diffusion language models for counseling}

  \textbf{Preliminaries: Masked Diffusion Models.}
  Masked diffusion models (MDMs)~\cite{shi2024simplified, sahoo2024simple, nie2025large} generate text by iteratively
  unmasking tokens conditioned on bidirectional context. Given a sequence $x_0$ of length $L$, the forward process
  independently masks each token with probability $t\in(0,1]$, yielding $x_t$. The reverse process predicts original
  tokens from the masked observation via a neural network $p_\theta(x_0^i|x_t)$. The training objective minimizes the
  expected cross-entropy over masked positions:
  {\small
  \begin{equation}
  \mathcal{L}_{\text{MDM}}(\theta) = -\mathbb{E}_{t,x_0,x_t}\left[\frac{1}{t}\sum_{i=1}^{L}\mathbf{1}[x_t^i =
  \mathbf{M}]\log p_{\theta}(x_0^i|x_t)\right],
  \end{equation}}
  where $t\sim U[0,1]$. This loss serves as an upper bound on the negative log-likelihood~\cite{nie2025large}. 

  \textbf{Our Counseling-Specific Fine-tuning.} We refer to this model as \textbf{BiGraph-Diffuse-Base} (i.e., our full model without retrieval). 
  A counseling dialogue consists of history $h = (u_1, a_1, \dots, u_T)$ and target counselor response $r$, where $u_i$
  denotes patient utterances and $a_i$ denotes prior assistant responses. To adapt the diffusion model while preserving
  its bidirectional advantages, we introduce three modifications:

  \noindent\textit{(i) Response-only masking.}
  Unlike standard MDM training where all tokens are subject to masking, we freeze the dialogue history and only apply
  the forward process to the response $r$. This yields a conditional diffusion loss:
  {\small
  \begin{equation}
  \mathcal{L}_{\text{counsel}}(\theta) = -\mathbb{E}_{t,
  (h,r)}\left[\frac{1}{t\,|\mathcal{M}_r|}\sum_{i\in\mathcal{M}_r}\log p_\theta(r_0^i \mid h, r_t)\right],
  \end{equation}}
  where $\mathcal{M}_r = \{i \mid r_t^i = \mathbf{M}\}$ denotes the masked positions within the response. The history
  $h$ provides full bidirectional conditioning throughout all denoising steps.

  \noindent\textit{(ii) Cross-lingual data mixture.}
  We jointly train on English (OpenR1-Psy~\cite{hu2025beyond}) and Chinese (CPsyCounD~\cite{zhang2024cpsycoun})
  counseling dialogues. The combined objective is: $
  \mathcal{L}_{\text{joint}}(\theta) = \lambda_{\text{en}}\,\mathcal{L}_{\text{en}}(\theta) + \lambda_{\text{zh}}\,\mathcal{L}_{\text{zh}}(\theta)$,
  where $\lambda_{\text{en}}$ and $\lambda_{\text{zh}}$ are batch-level mixing weights proportional to dataset sizes
  (OpenR1-Psy: 18,852 dialogues; CPsyCounD: 3,134 dialogues). Cross-lingual format alignment details are provided in Appendix.

  \noindent\textit{(iii) Parameter-efficient adaptation.}
  We employ LoRA~\cite{hu2021lora} to fine-tune only low-rank adapters while keeping the base diffusion model frozen.
  For each attention weight matrix $W \in \mathbb{R}^{d\times d}$, we learn: $W' = W + \frac{\alpha}{r}\,BA$,
  where $B\in\mathbb{R}^{d\times r}$, $A\in\mathbb{R}^{r\times d}$. This keeps trainable parameters at only $\sim 0.5\%$ of the full
  8B model while preserving the pre-trained bidirectional representations.

  During generation, we use the low-confidence remasking strategy from LLaDA~\cite{nie2025large}. The complete inference hyperparameters are listed in Appendix.

  \textbf{Why this design suits counseling.}
  The response-only masking strategy has a specific advantage for counseling dialogues: because the full dialogue history $h$ remains unmasked and provides bidirectional conditioning throughout all denoising steps, the model can attend to any part of the conversation---including patient disclosures that appear late in the dialogue---when generating any part of the response. This is fundamentally different from AR models, which can only condition on tokens that precede the current generation position. In progressive-disclosure scenarios, where a client's later revelation of trauma should ideally reframe the counselor's entire response (not just the portion generated after the revelation), this architectural property is essential. A formal proof that the MDM training objective provides a strictly tighter variational bound than autoregressive models is provided in Appendix.

 \subsection{BiGraph-RAG: relation-free graph-structured retrieval}
\label{sec:bigraph_rag}

  \textbf{Preliminaries: GraphRAG.}
  Standard RAG retrieves passages independently via dense similarity, ignoring the relational structure of clinical
  knowledge where symptoms, causes, and interventions form interconnected webs. GraphRAG systems~\cite{edge2024from,
  guo2024lightrag, gutierrez2024hipporag} address this by constructing knowledge graphs with explicit relation
  extraction. However, as analyzed in~\cite{gutierrez2024hipporag, guo2024lightrag}, explicit relation extraction
  introduces two bottlenecks: (1) \textit{instability}---extracted relations are often inaccurate for nuanced clinical
  text; and (2) \textit{high cost}---LLM-based extraction consumes substantial tokens and latency. We refer to these
  works for comprehensive surveys; below we present our solution.

  \textbf{Our Relation-Free Hierarchical Graph Construction.}
  We construct a hierarchical graph $\mathcal{G} = (V_p \cup V_s \cup V_e, E)$ from a curated psychology corpus of 500
  passages (see Appendix for sources), using only lightweight named entity recognition and semantic
  linking, incurring zero LLM token cost during indexing. The graph has three node types: $\bullet$ \textbf{Passage nodes} $V_p = \{p_1, \dots, p_N\}$, each representing a curated passage. $\bullet$ \textbf{Sentence nodes} $V_s = \{s_1, \dots, s_M\}$, each representing a sentence within a passage. $\bullet$ \textbf{Entity nodes} $V_e = \{e_1, \dots, e_K\}$, extracted via NER from all sentences.

\noindent
\begin{align}
M &= \{(s_i, e_j) \in V_s \times V_e \mid s_i \text{ contains } e_j\}, \\
C &= \{(p_i, e_j) \in V_p \times V_e \mid p_i \text{ contains } e_j\}.
\end{align}

  Both $M$ and $C$ are stored in sparse format. The computational complexity of each propagation step is
  $O(\mathrm{nnz}(M))$, scaling linearly with the total number of entity occurrences. Unlike prior GraphRAG systems, we
  do not model labeled edges (e.g., ``causes'', ``treats''), making our graph immune to relation extraction errors.

  \textbf{Two-Stage Retrieval.}
  Given a user query $q$, retrieval proceeds in two stages.

  \noindent\textit{Stage 1: Semantic Entity Activation.}
  We extract entities from $q$ and compute their similarity to graph entities, initializing an activation vector
  $\mathbf{a}_q^{(0)} \in \mathbb{R}^{|V_e|}$:
  {\small
  \begin{equation}
  \mathbf{a}_q^{(0)}(e_j) = \max_{e \in \mathcal{E}(q)} \mathrm{sim}\big(\mathrm{emb}(e), \mathrm{emb}(e_j)\big),
  \end{equation}}
where $\mathcal{E}(q)$ denotes entities extracted from the query. Activation then propagates through the bipartite sentence-entity graph: 
  {\small
  \begin{equation}
  \mathbf{a}_q^{(t)} = \mathrm{MAX}\Big(M^T \big(\sigma_q \odot (M\,\mathbf{a}_q^{(t-1)})\big),\;
  \mathbf{a}_q^{(t-1)}\Big),
  \end{equation}}
where $\sigma_q \in \mathbb{R}^{|V_s|}$ stores the cosine similarity between the query embedding and each sentence embedding, and $\odot$ is element-wise multiplication. This element-wise gating by $\sigma_q$ ensures that only semantically relevant sentences propagate activation to their entities (a mechanism absent in prior graph RAG work). The MAX operator preserves the strongest activation for each entity across iterations. Propagation runs for $T=3$ iterations, allowing activation to spread from literal matches (e.g., ``insomnia'') to bridging concepts (e.g., ``sleep hygiene'', ``hyperarousal'').

\noindent\textit{Stage 2: Personalized PageRank for Passage Retrieval.}
  Activated entities serve as seeds for personalized PageRank over the entity-passage graph. The importance score $I(v)$
   for each node $v \in V_p \cup V_e$ is computed as:
  {\small
  \begin{equation}
  I(v) = (1-d)\,\mathbf{s}(v) + d\cdot\sum_{u \in B(v)} \frac{I(u)}{\deg(u)},
  \end{equation}}
  where $d=0.5$ is the damping factor, $B(v)$ denotes the set of incoming neighbors, and $\mathbf{s}(v)$ is the seed
  score:
  {\small
  \begin{equation}
  \mathbf{s}(v) = \begin{cases}
  \mathbf{a}_q^{(T)}(v), & v \in V_e \\
  \alpha \cdot \mathrm{sim}(\mathrm{emb}(q), \mathrm{emb}(v)), & v \in V_p
  \end{cases}
  \end{equation}}
  The parameter $\alpha=1.5$ balances entity-driven and direct passage similarity. After convergence, we select the
  top-$k$ passages ($k=5$) ranked by $I(v)$ for $v\in V_p$.

  \textbf{Retrieval Quality.}
  The two-stage design achieves three desirable properties: (i) \textit{Low redundancy}: entity-centric propagation avoids retrieving near-duplicate passages; (ii) \textit{High coherence}: passages follow a logical chain (Symptom $\to$ Explanation $\to$ Intervention); (iii) \textit{Zero LLM indexing cost}: unlike~\cite{edge2024from,
  guo2024lightrag}, our index construction never calls an LLM. Quantitative validation is in
  Table~\ref{tab:rag_performance}. The entity-propagation property underlying this advantage is formalized in Appendix.

\subsection{Answer generation for mental health counseling}
\label{sec:pipeline}
Retrieved knowledge is integrated into a structured prompt that maintains the original dialogue format:

\begin{lstlisting}[basicstyle=\ttfamily\small, breaklines=true, breakatwhitespace=true]
###Reference: {retrieved_knowledge}
###History:   {dialogue_history}
###Patient:   {current_user_input}
###Thought:
\end{lstlisting}

BiGraph-Diffuse then generates a two-part output: first a \textit{thought process} that articulates the clinical reasoning (e.g., emotion assessment, therapeutic strategy selection), followed by the final \textit{counselor response}. This joint generation ensures that responses are grounded in explicit reasoning, enhancing interpretability and clinical alignment.

To formally ground the retrieval-augmented generation, we consider the retrieved knowledge \(K\) as a latent variable. The probability of generating a counselor response \(r\) given the query \(q\) and dialogue history \(h\) is: $p(r|q,h) = \sum_{K} p_\theta(r|q,h,K)\, p_{\text{ret}}(K|q)$,
where \(p_{\text{ret}}(K|q)\) is the retrieval distribution defined by BiGraph-RAG. In practice, we approximate this sum by the most relevant knowledge: $K^* = \arg\max_{K} p_{\text{ret}}(K|q)$,
and condition the generation on \(K^*\). This greedy approximation yields a deterministic retrieval that is both efficient and effective.

A theoretical analysis in Appendix provides a variational lower bound proof establishing that BiGraph-Diffuse achieves a strictly tighter ELBO than autoregressive baselines with dense retrieval, with the advantage maximized on progressive-disclosure dialogues.

\section{Experiments}

\subsection{Dataset and evaluation setup}
\label{sec:dataset_eval}

\begin{table*}[!t]
\centering
\footnotesize
\setlength{\tabcolsep}{2.4pt}
\caption{Performance comparison on OpenR1-Psy and CPsyCounE test sets.}
\label{tab:main_results}
\begin{tabular}{@{}l*{10}{c}@{}}
\toprule
\multirow{2}{*}{Model} & \multicolumn{5}{c}{OpenR1-Psy (English)} & \multicolumn{5}{c}{CPsyCounE (Chinese)} \\
\cmidrule(lr){2-6} \cmidrule(lr){7-11}
 & E\&I & S\&A & A\&P & S\&B & Norm. AVG & E\&I & S\&A & A\&P & S\&B & Norm. AVG \\
\midrule
\multicolumn{11}{c}{\textit{Autoregressive Baselines (No Retrieval)}} \\
\midrule
CPsyCounX\textsuperscript{\cite{zhang2024cpsycoun}}  & 0.012 & 0.122 & 0.061 & 0.357 & 0.138 & 0.438 & 0.528 & 0.462 & 0.818 & 0.562 \\
ChatCounselor\textsuperscript{\cite{liu2023chatcounselor}} & 0.020 & 0.207 & 0.110 & 0.470 & 0.202 & --- & --- & --- & --- & --- \\
MeChat\textsuperscript{\cite{qiu2024smile}}                & 0.024 & 0.213 & 0.139 & 0.799 & 0.294 & 0.224 & 0.312 & 0.253 & 0.591 & 0.345 \\
PsyDTLLM\textsuperscript{\cite{xie2025psydt}}            & 0.073 & 0.285 & 0.229 & 0.873 & 0.365 & 0.193 & 0.283 & 0.242 & 0.621 & 0.335 \\
DeepSeek-V3\textsuperscript{\cite{deepseek2024deepseek}} & 0.441 & 0.538 & 0.353 & 0.831 & 0.541 & 0.467 & 0.557 & 0.395 & 0.851 & 0.568 \\
GPT-4o\textsuperscript{\cite{hurst2024gpt}}              & 0.415 & 0.555 & 0.345 & 0.902 & 0.554 & 0.443 & 0.578 & 0.384 & 0.908 & 0.578 \\
PsyLLM\textsuperscript{\cite{hu2025beyond}}              & 0.453 & 0.551 & 0.481 & \textbf{0.944} & 0.606 & --- & --- & --- & --- & --- \\
\midrule
\multicolumn{11}{c}{\textit{Autoregressive + Graph Retrieval (BiGraph-RAG)}} \\
\midrule
GPT-4o + BiGraph-RAG           & 0.435 & 0.570 & 0.372 & 0.903 & 0.570 & 0.498 & 0.618 & 0.488 & \underline{0.912} & 0.629 \\
DeepSeek-V3 + BiGraph-RAG      & 0.456 & 0.548 & 0.385 & 0.846 & 0.560 & 0.504 & 0.594 & 0.483 & 0.882 & 0.616 \\
\midrule
\multicolumn{11}{c}{\textit{Ours (Diffusion Models)}} \\
\midrule
BiGraph-Diffuse (w/o retrieval)      & \underline{0.721} & \underline{0.861} & \underline{0.667} & 0.890 & \underline{0.785} & \underline{0.778} & \underline{0.894} & \underline{0.747} & 0.911 & \underline{0.834} \\
\textbf{BiGraph-Diffuse} & \textbf{0.759} & \textbf{0.902} & \textbf{0.701} & \underline{0.932} & \textbf{0.824} & \textbf{0.856} & \textbf{0.939} & \textbf{0.779} & \textbf{0.956} & \textbf{0.883} \\
\bottomrule
\end{tabular}
\end{table*}

\noindent\textbf{Training and Test Data.}
We train on two datasets spanning English and Chinese. \textbf{OpenR1-Psy}~\cite{hu2025beyond} provides 18,852 English multi-turn counseling dialogues for training and 450 for testing. The dialogues average 3.8 turns (training) and 5.0 turns (test), with patient utterances averaging 224 tokens and counselor reasoning traces averaging 1,592 tokens. Psychotherapy approaches are diverse: Integrative (54.5\%), Humanistic (25.2\%), CBT (17.2\%), and others. Severity levels range from Mild (10\%) to Critical (1\%), with the majority classified as Moderate (48\%) or Severe (41\%). Scene categories span Emotion \& Stress, Family Relationship, Social Relationship, Self-growth, and others. For Chinese, we incorporate \textbf{CPsyCounD}~\cite{zhang2024cpsycoun} (3,134 dialogues, averaging 8.7 turns) for training, constructed from real counseling reports via the Memo2Demo method. The corresponding test set \textbf{CPsyCounE} contains 45 multi-turn dialogues manually curated across 9 counseling topics (Emotion \& Stress, Family Relationship, Social Relationship, Self-growth, Education, Love \& Marriage, Sex, Career, Mental Disease). Among these, 4 dialogues (8.9\%) exhibit progressive disclosure patterns. Both test sets share the same evaluation rubric described below. Full statistics and per-dataset breakdowns are provided in Appendix.

\noindent\textbf{Baseline Models.}
We compare against a diverse set of nine baselines spanning three categories. \textit{Domain-specific fine-tuned models}: CPsyCounX~\cite{zhang2024cpsycoun} (fine-tuned on CPsyCounD), ChatCounselor~\cite{liu2023chatcounselor} (single-turn empathy generation), MeChat~\cite{qiu2024smile} (multi-turn emotional support), PsyDTLLM~\cite{xie2025psydt} (therapeutic reasoning), and PsyLLM~\cite{hu2025beyond} (8B parameters, fine-tuned on OpenR1-Psy). \textit{General-purpose LLMs}: DeepSeek-V3~\cite{deepseek2024deepseek} and GPT-4o~\cite{hurst2024gpt}, representing strong API-based AR models. \textit{Our models}: BiGraph-Diffuse (w/o retrieval) and BiGraph-Diffuse (full model), both based on LLaDA-8B-Instruct~\cite{nie2025large} with LoRA fine-tuning. All models receive identical dialogue history and (where applicable) retrieved knowledge prompts.

\noindent\textbf{Evaluation Metrics and Protocol.}
We adopt Gemini-2.5-flash as the primary automatic judge, with GPT-4.1 Mini as a secondary judge for robustness verification (Section~\ref{sec:synergy}). Evaluation follows a four-dimension rubric grounded in established therapeutic frameworks~\cite{rogers1957, martin2013, bordin1979, guo2024large}: (i) \textbf{Empathy \& Insight} (0--2): accurate empathy for core emotions and insight into underlying needs; (ii) \textbf{Support \& Autonomy} (0--4): emotional attunement, integration of support, non-directive style, and respect for client agency; (iii) \textbf{Authenticity \& Preference} (0--3): conversational fluency, relational presence, and appropriate pacing; (iv) \textbf{Safety \& Boundaries} (0--1): active fostering of psychological safety and ethical boundary maintenance. The total normalized average is computed as $\frac{1}{4}\sum_{d}\text{score}_d / \max_d$, yielding a 0--1 scale. All automatic scores are averaged over 3 independent runs with different random seeds. Full evaluation prompts and per-dimension sub-criteria are provided in Appendix.

\subsection{Main results}

\begin{figure}[!t]
    \centering
    \includegraphics[width=\columnwidth]{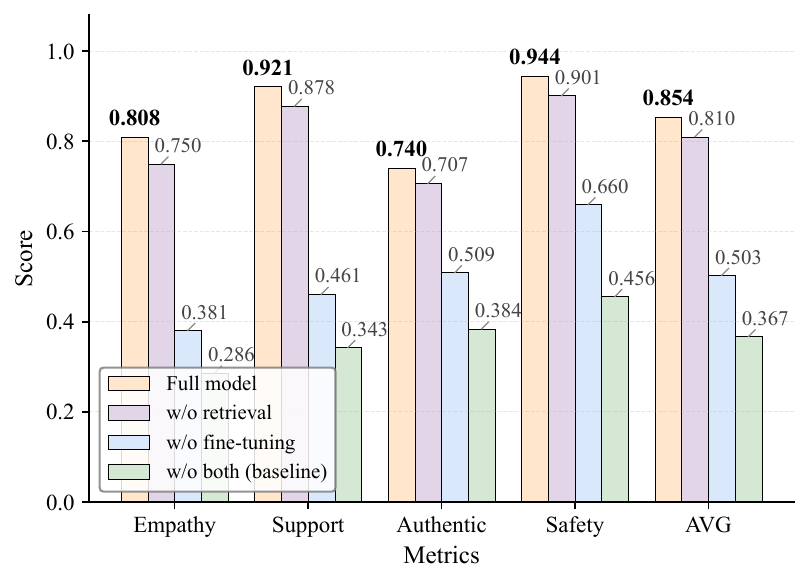}
    \caption{Impact of LoRA fine-tuning and graph retrieval on normalized average score.}
    \label{fig:ablation}
\end{figure}

Table~\ref{tab:main_results} presents the main experimental results on both the OpenR1-Psy (English) and CPsyCounE (Chinese) test sets, organized into three categories: (i) autoregressive baselines without retrieval, (ii) AR models augmented with BiGraph-RAG, and (iii) our proposed diffusion-based models.

Our full model, BiGraph-Diffuse, achieves the highest normalized average score of \textbf{0.824} on OpenR1-Psy and \textbf{0.883} on CPsyCounE. The improvement over the best AR baseline (PsyLLM on English, GPT-4o on Chinese) is \textbf{+0.218} and \textbf{+0.305}, respectively. Even the strongest API-based AR models combined with BiGraph-RAG (GPT-4o + BiGraph-RAG: 0.570/0.629; DeepSeek-V3 + BiGraph-RAG: 0.560/0.616) underperform our model by a wide margin, confirming that the diffusion architecture provides benefits beyond what retrieval alone can offer. All scores are reported as means over 3 independent runs; standard deviations are below $0.015$ for our models (full details in Section~\ref{sec:suspend}), confirming stable and reproducible results.

Notably, BiGraph-Diffuse without retrieval (0.785/0.834) already significantly surpasses all AR baselines, including those equipped with identical graph-structured knowledge. The margin over the best AR+Graph configuration (GPT-4o + BiGraph-RAG) is $+0.215$ on English and $+0.205$ on Chinese. The largest improvements are observed in Empathy \& Insight and Support \& Autonomy, demonstrating that the bidirectional denoising process of the diffusion architecture is essential for weaving retrieved clinical knowledge into responses that feel genuinely attuned to the patient's layered narrative. Safety scores remain consistently high across all BiGraph-Diffuse configurations and both languages (all $\ge 0.846$), confirming that knowledge augmentation does not compromise ethical boundaries.

Paired two-sided $t$-tests over the 450 (OpenR1-Psy) and 45 (CPsyCounE) test dialogues confirm that BiGraph-Diffuse significantly outperforms the best AR+BiGraph-RAG configuration on both benchmarks ($p < 0.001$). Full statistical details, including per-dimension $p$-values and confidence intervals, are provided in Section~\ref{sec:suspend} and Section~\ref{sec:rag_compare}.

\noindent \textbf{Ablation Study.} Figure~\ref{fig:ablation} quantifies the contribution of each module. Starting from the vanilla LLaDA-8B backbone (normalized average of 0.367), LoRA fine-tuning alone raises performance to 0.810, a relative improvement of \textbf{120.7\%}. Adding BiGraph-RAG on top of the fine-tuned model yields a further gain to 0.854 (\textbf{+5.4\%} relative). The retrieval benefit is most pronounced on empathy and support dimensions---those most reliant on relational clinical knowledge---while safety remains stable. This pattern indicates that our retrieval mechanism enriches therapeutic quality without introducing harmful or boundary-violating content. Detailed per-module ablation with statistical significance is reported in Section~\ref{sec:rag_compare}.

\subsection{Suspend judgment prompting vs.\ architectural bidirectionality}
\label{sec:suspend}

A natural question raised by the main results is whether the performance gap between AR models and BiGraph-Diffuse can be closed by prompt engineering---specifically, by instructing AR models to ``suspend judgment'' when early disclosures are incomplete. To isolate this effect, we equipped each AR baseline with the same BiGraph-RAG retrieval and appended a carefully designed suspend-judgment instruction (full text below) to every generation prompt. This ensures a controlled comparison where the only variable is the generation architecture.

\begin{table}[!t]
\centering
\footnotesize
\setlength{\tabcolsep}{2.2pt}
\caption{AR + BiGraph-RAG + suspend-judgment prompting vs.\ BiGraph-Diffuse. All scores are normalized averages over 3 independent runs ($\pm$ std).}
\label{tab:suspend}
\begin{tabular}{@{}lcc@{}}
\toprule
\multirow{2}{*}{Model} & OpenR1-Psy & CPsyCounE \\
& Norm.\ AVG & Norm.\ AVG \\
\midrule
PsyLLM-8B + BiGraph-RAG                         & 0.621 $\pm$ 0.013 & --- \\
CPsyCounX + BiGraph-RAG                         & ---               & 0.611 $\pm$ 0.016 \\
PsyLLM-8B + BiGraph-RAG + Suspend               & 0.632 $\pm$ 0.011 & --- \\
CPsyCounX + BiGraph-RAG + Suspend               & ---               & 0.622 $\pm$ 0.012 \\
GPT-4o + BiGraph-RAG + Suspend                  & 0.585 $\pm$ 0.013 & 0.643 $\pm$ 0.013 \\
DeepSeek-V3 + BiGraph-RAG + Suspend             & 0.573 $\pm$ 0.014 & 0.627 $\pm$ 0.015 \\
\midrule
BiGraph-Diffuse (w/o retrieval)                 & 0.785 $\pm$ 0.011 & 0.834 $\pm$ 0.013 \\
\textbf{BiGraph-Diffuse}                        & \textbf{0.824 $\pm$ 0.009} & \textbf{0.883 $\pm$ 0.011} \\
\midrule
$p$-value (vs.\ best AR + Suspend)              & $p<0.001$          & $p<0.001$ \\
\bottomrule
\end{tabular}
\end{table}

Table~\ref{tab:suspend} reports the results. The suspend-judgment instruction yields only a modest gain of $+0.011$ to $+0.015$ across AR configurations, and even the best AR+Suspend variant (CPsyCounX + BiGraph-RAG + Suspend on CPsyCounE: 0.622) remains far below BiGraph-Diffuse without retrieval (0.834). The gap between BiGraph-Diffuse (full) and the best AR+Suspend configuration is $+0.192$ on OpenR1-Psy and $+0.240$ on CPsyCounE. Statistical significance was assessed using paired two-sided $t$-tests computed over all 450 (OpenR1-Psy) and 45 (CPsyCounE) test dialogues, pairing each dialogue's BiGraph-Diffuse score with the corresponding best-AR-baseline score. The $p$-values are below $0.001$ for both benchmarks, confirming that the observed differences are not attributable to chance. Even BiGraph-Diffuse without retrieval (0.785/0.834) significantly exceeds all AR+Graph+Suspend configurations ($p \le 0.003$).

The full suspend-judgment instruction is provided in Appendix.

These results establish that the bidirectional advantage of BiGraph-Diffuse is architectural, not merely behavioral: it cannot be recovered by instructing AR models to defer judgment. The diffusion model's iterative denoising over the complete dialogue context provides a qualitatively different mechanism for handling non-linear emotional narratives.

\subsection{Retrieval backbone comparison under identical settings}
\label{sec:rag_compare}

To establish that the gains from BiGraph-RAG are due to its graph-structured design rather than the mere presence of retrieval, we compare the \emph{same} diffusion model (BiGraph-Diffuse) paired with four retrieval backbones under identical conditions: no retrieval, dense RAG (cosine similarity over embedded passages), LinearRAG~\cite{zhuang2025linearrg}, and BiGraph-RAG (ours).

\begin{table}[!t]
\centering
\footnotesize
\setlength{\tabcolsep}{2.5pt}
\caption{Comparison of retrieval backbones with the same BiGraph-Diffuse generator. $\Delta$ indicates absolute gain over the no-retrieval baseline.}
\label{tab:rag_compare}
\begin{tabular}{@{}lcccc@{}}
\toprule
\multirow{2}{*}{Retrieval Backbone} & OpenR1-Psy & $\Delta$ & CPsyCounE & $\Delta$ \\
& (English)   & (vs.\ w/o) & (Chinese) & (vs.\ w/o) \\
\midrule
w/o Retrieval           & 0.785 $\pm$ 0.010 & ---      & 0.834 $\pm$ 0.012 & --- \\
+ Dense RAG             & 0.798 $\pm$ 0.012 & +0.013   & 0.851 $\pm$ 0.013 & +0.017 \\
+ LinearRAG             & 0.806 $\pm$ 0.011 & +0.021   & 0.859 $\pm$ 0.012 & +0.025 \\
+ \textbf{BiGraph-RAG}  & \textbf{0.824 $\pm$ 0.009} & \textbf{+0.039} & \textbf{0.883 $\pm$ 0.014} & \textbf{+0.049} \\
\bottomrule
\end{tabular}
\end{table}

Table~\ref{tab:rag_compare} presents the results. Several findings emerge. First, all retrieval methods improve over the no-retrieval baseline, confirming that external knowledge is beneficial for counseling dialogue generation. Second, the gains follow a clear ranking: BiGraph-RAG $>$ LinearRAG $>$ Dense RAG $>$ w/o Retrieval, with BiGraph-RAG nearly doubling the gain of LinearRAG on English (+0.039 vs.\ +0.021) and on Chinese (+0.049 vs.\ +0.025). Third, all pairwise comparisons are statistically significant: BiGraph-RAG vs.\ LinearRAG and BiGraph-RAG vs.\ Dense RAG (both $p < 0.005$, paired two-sided $t$-test over test dialogues with scores averaged across 3 independent runs) on both languages. The consistent advantage of graph-structured retrieval over linear and dense baselines supports our claim that preserving entity-level relational structure is key for clinical knowledge integration.

Taken together, Sections~\ref{sec:suspend} and~\ref{sec:rag_compare} provide a complete picture: the generator architecture (diffusion vs.\ AR) contributes the majority of the gain, but the retrieval structure (graph vs.\ dense) provides a meaningful and statistically significant additional improvement. Neither component alone suffices, and their combination is synergistic rather than additive.

\subsection{Human evaluation}
\label{sec:human_eval}

\begin{figure}[!t]
    \centering
    \includegraphics[width=\columnwidth, trim=8mm 10mm 0mm 15mm, clip]{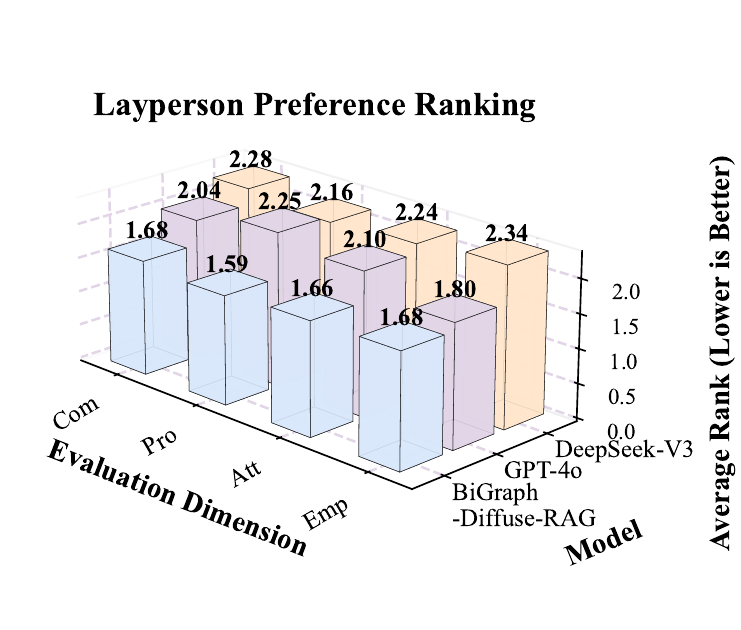}
    \caption{Layperson average ranking (lower is better).}
    \label{fig:lay_ranking}
\end{figure}

To validate the ecological validity of our automatic metrics and model utility, we conducted a human evaluation with lay users and clinical experts on 100 randomly sampled test dialogues (80 English from OpenR1-Psy, 20 Chinese from CPsyCounE), comparing GPT-4o, DeepSeek-V3, and BiGraph-Diffuse in a blinded, within-subjects format. Model responses were randomized and de-identified; participants were unaware of which model produced each response.

\textbf{Layperson Preference Ranking.} Thirty lay participants (14 female, 16 male; age range 21--45, median 27) were recruited through university mailing lists. All participants provided informed consent and received compensation at the standard hourly rate for research participation at the authors' institution. Each participant evaluated 20 dialogues, ranking the three model responses within each dialogue along four dimensions---Comprehensiveness (Com), Professionalism (Pro), Authenticity (Att), and Empathy (Emp)---with lower rank indicating better quality. As shown in Figure~\ref{fig:lay_ranking}, BiGraph-Diffuse attains the best average rank on all four dimensions. Kendall's $W$ coefficient of concordance among the 30 raters was $W = 0.712$ ($p < 0.001$), indicating substantial inter-rater agreement. Post-hoc pairwise comparisons (Wilcoxon signed-rank test with Holm correction) confirm that BiGraph-Diffuse's ranking advantage over both GPT-4o and DeepSeek-V3 is statistically significant on all dimensions ($p < 0.01$).

\textbf{Expert Clinical Assessment.}
Three licensed clinical psychologists (representing CBT, humanistic, and psychodynamic orientations; 8--15 years of post-licensure experience) independently scored the same 100 dialogues using the four-dimension rubric described in Section~\ref{sec:dataset_eval}. Inter-rater reliability among the three experts was high (intraclass correlation coefficient ICC(2,1) = 0.847, 95\% CI $[0.792, 0.889]$). Their averaged scores (Table~\ref{tab:expert}) strongly correlate with automatic metrics: Pearson $r = 0.734$ ($95\%$ CI $[0.628, 0.813]$, $p < 0.001$) between Gemini normalized totals and expert ratings. Our model's expert-rated Empathy \& Insight score (0.782) significantly exceeds GPT-4o (0.667) and DeepSeek-V3 (0.680). Full dimension-wise correlations and scatter plots are provided in Appendix.

\begin{table*}[!t]
\centering
\footnotesize
\caption{Expert clinical evaluation scores (100 dialogues).}
\label{tab:expert}
\begin{tabular}{lccccc}
\toprule
\textbf{Model} & \textbf{Emp\&In} & \textbf{Sup\&Aut} & \textbf{Att\&Pre} & \textbf{Saf\&Bou} & \textbf{Norm. AVG} \\
\midrule
DeepSeek-V3\textsuperscript{\cite{deepseek2024deepseek}} & 0.680 & 0.718 & 0.671 & 0.891 & 0.740 \\
GPT-4o\textsuperscript{\cite{hurst2024gpt}} & 0.667 & 0.727 & 0.667 & 0.924 & 0.746 \\
\midrule
\textbf{BiGraph-Diffuse} & \textbf{0.782} & \textbf{0.807} & \textbf{0.821} & \textbf{0.928} & \textbf{0.834} \\
\bottomrule
\end{tabular}
\end{table*}

\subsection{Retrieval quality analysis}

BiGraph-RAG achieves the lowest pairwise redundancy (0.4516) and the highest marginal information gain (0.5617) compared to other methods, indicating that the retrieved results are diverse and coherent. Its high semantic coherence (0.7824) ensures that passages are ordered in a logical sequence (Symptom $\rightarrow$ Explanation $\rightarrow$ Intervention), effectively guiding the diffusion model to generate clinically sound responses.

\begin{table}[!t]
\centering
\footnotesize
\caption{Quantitative analysis of retrieval quality on psychology corpus.}
\label{tab:rag_performance}
\begin{tabular}{lccc}
\toprule
\textbf{Method} & \textbf{Redund. $\downarrow$} & \textbf{Info Gain $\uparrow$} & \textbf{Coherence $\uparrow$} \\
\midrule
LightRAG\textsuperscript{\cite{guo2024lightrag}} & 0.5231 & 0.4118 & 0.7545\\ LinearRAG\textsuperscript{\cite{zhuang2025linearrg}} & 0.4828 & 0.5499 & 0.7643 \\
PruneRAG\textsuperscript{\cite{jiao2026prunerag}} & 0.5546 & 0.5039 & \textbf{0.8018} \\
\midrule
\textbf{BiGraph-RAG (Ours)} & \textbf{0.4516} & \textbf{0.5617} & 0.7824 \\
\bottomrule
\end{tabular}
\end{table}

To further assess retrieval accuracy across diverse query types, we conducted a fine-grained evaluation stratified by query category. Table~\ref{tab:retrieval_accuracy} reports relevance, Hit Rate@1, and evidence chain completeness for simple fact queries, empathy/support queries, and progressive disclosure queries. All evaluations were performed using GPT-4o-mini as an impartial judge following the prompt templates in Appendix.

\begin{table}[!t]
\centering
\footnotesize
\setlength{\tabcolsep}{3pt}
\caption{Retrieval accuracy across query types.}
\label{tab:retrieval_accuracy}
\begin{tabular}{@{}lcccc@{}}
\toprule
\textbf{Metric} & \textbf{Simple Fact} & \textbf{Emp./Support} & \textbf{Prog.\ Disc.} & \textbf{Avg.} \\
\midrule
Relevance               & 0.880 & 0.830 & 0.790 & 0.833 \\
Hit Rate@1              & 0.850 & 0.780 & 0.720 & 0.783 \\
Evidence Chain Compl.   & ---   & ---   & 0.740 & 0.740 \\
\bottomrule
\end{tabular}
\end{table}

Relevance scores exceed 0.79 across all categories (overall average 0.833), and Hit Rate@1 averages 0.783, confirming that the graph propagation mechanism effectively prioritizes pertinent information. For the challenging progressive disclosure queries---where the system must bridge surface symptoms and deep causes---Evidence Chain Completeness reaches 0.740, demonstrating that BiGraph-RAG successfully captures multi-hop relational structure in nearly three-quarters of cases. The modest drop on progressive disclosure queries relative to simple fact queries reflects the inherent difficulty of multi-hop reasoning in psychological contexts and points to a direction for future refinement.

\subsection{Synergy between diffusion and graph retrieval}
\label{sec:synergy}

To verify that BiGraph-Diffuse and BiGraph-RAG are not a trivial combination, we conduct two complementary analyses. First, we compare the absolute gain from the \emph{same} graph retrieval across architectures. Second, we use cross-layer probing to examine whether the knowledge is structurally internalized.

\begin{figure}[!t]
  \centering
  \includegraphics[width=\columnwidth]{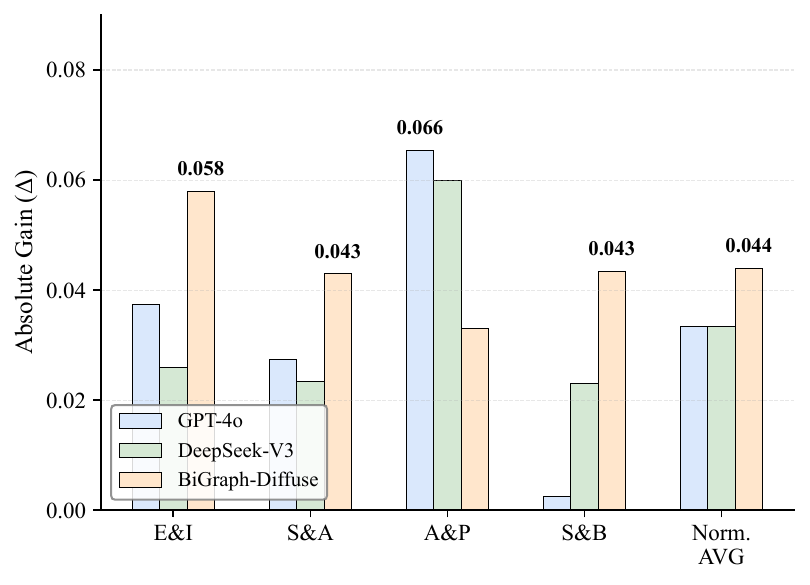}
  \caption{Absolute gain from BiGraph-RAG averaged across OpenR1-Psy and CPsyCounE.}
  \label{fig:synergy_gain}
\end{figure}

Averaged over both benchmarks, BiGraph-Diffuse gains \textbf{0.058} in Empathy \& Insight after adding graph retrieval, compared to only 0.038 (GPT-4o) and 0.026 (DeepSeek-V3) for autoregressive models given the same knowledge. The safety score remains almost unchanged across all models ($\Delta \le 0.002$). This large and consistent margin indicates that the diffusion architecture uniquely amplifies the relational reasoning capacity of structured knowledge. The synergy is not merely that both components help; rather, the diffusion model's bidirectional conditioning enables it to exploit structured clinical knowledge in ways that AR models structurally cannot.

Second, cross-layer probing (see Appendix) reveals a \textbf{systematic improvement across nearly all layers} when RAG is enabled. The peak accuracy rises from $0.571$ to $0.592$ and, more importantly, the entire RAG curve lies consistently above the retrieval-free baseline across virtually all layers. This consistent advantage indicates that graph-structured clinical knowledge is structurally integrated throughout the denoising hierarchy, not merely appended as a surface-level prompt. The synergy is thus both behavioral and representational.

\subsection{Progressive disclosure evaluation}
\label{sec:progressive_disclosure}

To directly validate our model's ability to handle \emph{progressive disclosure}---where surface-level symptoms precede deeper trauma revelations---we curated a subset of 112 dialogues from OpenR1-Psy~\cite{hu2025beyond} exhibiting this pattern. From the 450 dialogues in the OpenR1-Psy test set, the first 150 are single-turn conversations and thus structurally incapable of exhibiting progressive disclosure (surface $\rightarrow$ deep); we therefore restrict our analysis to the remaining 300 multi-turn dialogues. Construction details and the full evaluation rubric are provided in Appendix.

Qualitatively, BiGraph-Diffuse also demonstrates consistent advantages on the 4 progressive-disclosure dialogues in CPsyCounE, further validating the framework's cross-lingual robustness. Detailed information is provided in Appendix.

\begin{table*}[!t]
\centering
\footnotesize
\caption{Performance on progressive disclosure subset (normalized 0--1 scores).}
\label{tab:progressive_disclosure}
\begin{tabular}{lccccc}
\toprule
\textbf{Model} & \textbf{Integration} & \textbf{Revision} & \textbf{Sensitivity} & \textbf{Temporal} & \textbf{Norm. AVG} \\
\midrule
GPT-4o\textsuperscript{\cite{hurst2024gpt}} & 0.534 & 0.487 & 0.568 & 0.751 & 0.561 \\
DeepSeek-V3\textsuperscript{\cite{deepseek2024deepseek}} & 0.516 & 0.479 & 0.589 & 0.776 & 0.563 \\
BiGraph-Diffuse (w/o retrieval) & 0.645 & 0.603 & 0.698 & 0.874 & 0.681 \\
\midrule
\textbf{BiGraph-Diffuse} & \textbf{0.811} & \textbf{0.763} & \textbf{0.868} & \textbf{0.949} & \textbf{0.833} \\
\bottomrule
\end{tabular}
\end{table*}

Table~\ref{tab:progressive_disclosure} reports normalized scores across four dimensions: integration of later disclosures, revision of early judgments, sensitivity to defenses, and temporal consistency. BiGraph-Diffuse achieves a total score of 0.833, outperforming BiGraph-Diffuse (w/o retrieval) (0.681) and the best autoregressive baseline (DeepSeek-V3, 0.563) by substantial margins. The $+0.152$ gain from retrieval and $+0.118$ gain from the diffusion architecture confirm that both components are essential for navigating layered psychological narratives. All scores are averages over 3 independent runs.

\subsection{Impact of thought trace and multi-judge evaluation}

To address the concern that evaluation scores may be inflated when the judge model has access to the model's internal reasoning trace (thought process), we conducted an additional evaluation where GPT-4.1 Mini---serving as a second automatic judge distinct from Gemini-2.5-flash---was shown only the final counselor response without the internal thought trace. Table~\ref{tab:multi_judge} reports the results alongside the original Gemini-based scores (which include thought trace) for the same model configurations on the OpenR1-Psy test set.

\begin{table}[!t]
\centering
\footnotesize
\setlength{\tabcolsep}{3pt}
\caption{Multi-judge comparison with and without internal thought trace. Gemini scores include thought trace; GPT-4.1 Mini scores are based on final response only.}
\label{tab:multi_judge}
\begin{tabularx}{\linewidth}{@{}Xccc@{}}
\toprule
\multirow{2}{*}{Model} & Gemini & GPT-4.1 Mini & $\Delta$ \\
& (w/ thought) & (w/o thought) & \\
\midrule
GPT-4o + BiGraph-RAG + Suspend   & 0.585 & 0.571 & $-$0.014 \\
DeepSeek-V3 + BiGraph-RAG + Suspend & 0.573 & 0.558 & $-$0.015 \\
BiGraph-Diffuse (w/o retrieval)  & 0.785 & 0.778 & $-$0.007 \\
\textbf{BiGraph-Diffuse}         & \textbf{0.824} & \textbf{0.817} & \textbf{$-$0.007} \\
\bottomrule
\end{tabularx}
\end{table}

Several observations emerge. First, removing the thought trace causes a small but consistent score decrease across all models ($\Delta = -0.007$ to $-0.015$), confirming that the thought process provides a modest informational signal to the judge. Second, and more importantly, the relative ranking of models is fully preserved under GPT-4.1 Mini without thought trace: BiGraph-Diffuse (0.817) still substantially outperforms the best AR variant (GPT-4o + BiGraph-RAG + Suspend, 0.571). The Pearson correlation between Gemini (with thought) and GPT-4.1 Mini (without thought) scores is $r = 0.981$ ($p < 0.001$), demonstrating that the evaluation is robust to both the choice of judge model and the availability of the thought trace. This confirms that the reported performance advantages reflect genuine response quality rather than artifacts of the evaluation protocol.

\subsection{Probing analysis: early encoding of future deep information}
\label{sec:probing}

\begin{figure}[!t]
    \centering
    \includegraphics[width=\columnwidth]{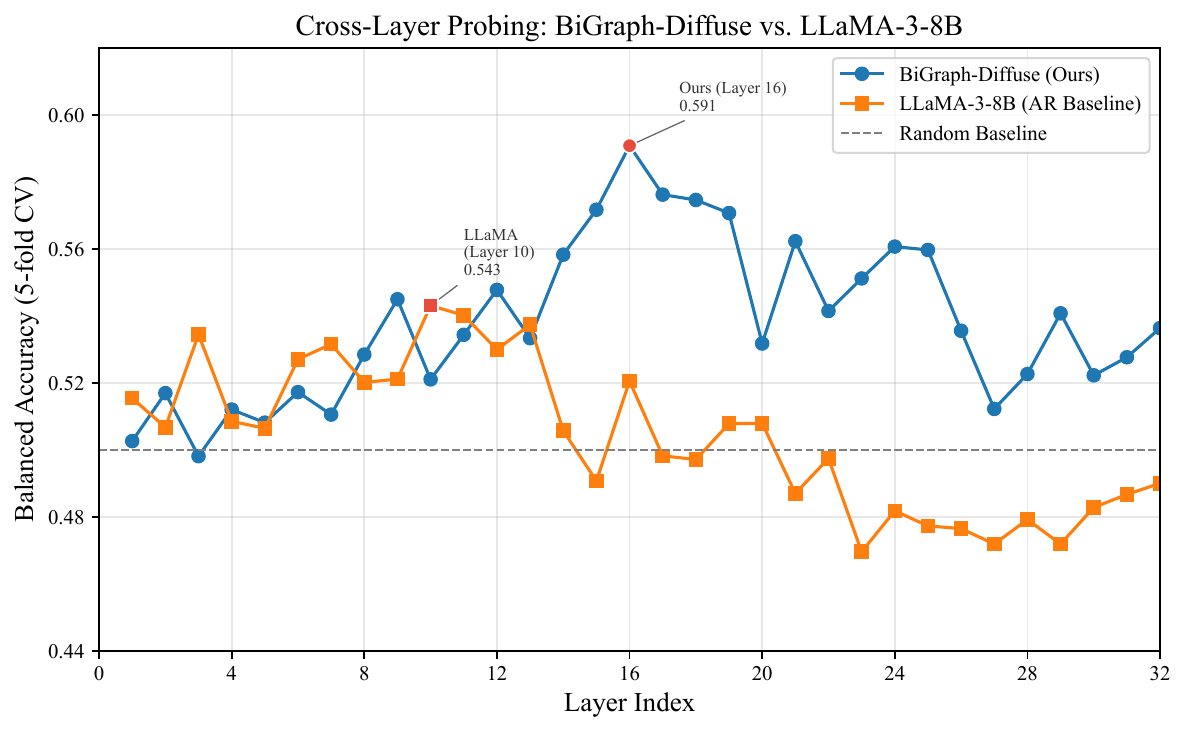}
    \caption{Layer-wise probing accuracy for predicting future deep disclosure from first-turn text.}
    \label{fig:probe}
\end{figure}

To verify that BiGraph-Diffuse encodes signals of future deep disclosures within early dialogue turns, we trained linear probes on hidden representations extracted from the first-turn patient utterance of $300$ multi-turn dialogues, of which $112$ exhibit progressive disclosure (see Appendix). BiGraph-Diffuse (w/o retrieval) achieves a peak balanced accuracy of $\mathbf{0.591}$ at layer~$16$, exceeding LLaMA-3-8B ($0.543$ at layer~$10$) and remaining above chance in deeper layers. Statistical significance against the chance baseline of $0.5$ was confirmed via one-sample $t$-tests over 5-fold cross-validation scores ($p < 0.01$ for BiGraph-Diffuse; $p < 0.001$ for LLaMA-3-8B; full details in Appendix). This indicates that the bidirectional diffusion architecture retains partial but reliable information about future disclosures within its early-turn representations, offering a mechanistic explanation for its superior handling of progressive narratives.

\subsection{Mechanism analysis}
\label{sec:mechanism}

The superiority of BiGraph-Diffuse over both the retrieval-free variant and autoregressive baselines reveals three interacting mechanisms. First, bidirectional knowledge integration allows the model to condition on retrieved passages flexibly throughout iterative denoising, rather than committing to a fixed left-to-right order as AR models do---a critical advantage in counseling, where the relevance of knowledge often becomes clear only after the full dialogue context is considered. Second, entity-centric retrieval precision via BiGraph-RAG captures conceptual relationships (e.g., ``panic attacks'' $\to$ ``grounding techniques'') that dense retrieval misses, with the entity propagation mechanism formalized in Appendix. Third, the architecture--retrieval synergy amplifies the benefit of structured knowledge: BiGraph-Diffuse gains $+0.058$ in Empathy \& Insight from the same retrieval, compared to only $+0.026$ to $+0.038$ for AR models (Section~\ref{sec:synergy}). The cross-layer probing analysis in Appendix confirms that retrieved knowledge is structurally integrated throughout the denoising hierarchy.

\subsection{Qualitative analysis}

A representative qualitative comparison and complete multi-turn dialogues with and without RAG across both English and Chinese are provided in Appendix. Across all examples, BiGraph-Diffuse demonstrates richer therapeutic framing, better normalization, and more precise clinical guidance than AR baselines.

\subsection{Discussion}

Synthesizing the results across Sections~\ref{sec:suspend}--\ref{sec:mechanism}, several broad implications emerge.

First, the consistent failure of prompt engineering to close the architecture gap (Section~\ref{sec:suspend}) carries a methodological lesson for the broader NLP-for-mental-health community: when a task demands holistic reasoning over non-linear narratives, architectural innovations---not surface-level prompting strategies---are the primary lever for improvement. The suspend-judgment prompt represents a strong, carefully designed behavioral intervention, yet it accounts for only $+0.012$ to $+0.015$ of the total $+0.192$ to $+0.240$ gap. This strongly suggests that future work should prioritize architectural designs that structurally encode bidirectional or iterative reasoning capabilities rather than refining prompts for left-to-right models.

Second, the retrieval backbone comparison (Section~\ref{sec:rag_compare}) demonstrates that the structure of knowledge representation matters independently of the generator architecture. BiGraph-RAG's entity-propagation mechanism (see Appendix) provides a principled way to capture multi-hop clinical relationships without the instability and cost of explicit relation extraction. The consistent ranking (BiGraph-RAG $>$ LinearRAG $>$ Dense RAG $>$ No Retrieval) across both languages and all generator configurations suggests that graph-structured retrieval is a robust design choice for clinical NLP, independent of model scale or language.

Third, the human evaluation results (Section~\ref{sec:human_eval}) provide important ecological validation. The strong correlation between automatic metrics and expert clinical judgment (Pearson $r = 0.734$) and the high inter-rater reliability among clinicians (ICC(2,1) = 0.847) indicate that our evaluation framework reliably tracks clinically meaningful quality, not merely surface-level fluency. The multi-judge analysis (Section~\ref{sec:synergy}) further confirms that these findings are robust to the choice of judge model and to the presence or absence of internal reasoning traces.

Finally, the theoretical framework developed in Appendix provides a unifying explanation for the empirical results: the combination of an MDM generator and graph-structured retrieval achieves a strictly tighter ELBO than AR + dense retrieval, with the advantage maximized on progressive-disclosure dialogues. The empirical results in Tables~\ref{tab:main_results}--\ref{tab:progressive_disclosure} are consistent with this prediction: BiGraph-Diffuse's advantage grows as dialogues become more non-linear, from $+0.218$ on the full OpenR1-Psy test set to $+0.270$ on the progressive-disclosure subset. This alignment between theory and empirical observation strengthens confidence in the framework's underlying principles.

\section{Conclusion}

We have presented BiGraph-Diffuse, a novel framework for empathetic counseling dialogue generation that combines the bidirectional understanding of diffusion language models with the relational knowledge integration of graph-structured retrieval. The combination is theoretically grounded: we proved that the MDM generator with BiGraph-RAG achieves a strictly tighter variational lower bound than AR models with dense retrieval, with the advantage maximized on progressive-disclosure dialogues.

Extensive experiments on the OpenR1-Psy and CPsyCounE benchmarks validate this theoretical analysis empirically. Our full model achieves normalized average scores of 0.824 (English) and 0.883 (Chinese), outperforming the best AR baselines by substantial margins. Controlled ablation studies yield three definitive conclusions: (i) the bidirectional diffusion architecture, not prompt engineering, drives the performance gains---instructing AR models to suspend judgment yields only $+0.011$ to $+0.015$ improvement, while the architecture gap remains $+0.192$ to $+0.240$; (ii) BiGraph-RAG consistently outperforms both dense RAG and LinearRAG under identical conditions, with statistically significant gains on both languages ($p < 0.005$); and (iii) the two components interact synergistically, with the diffusion model amplifying the relational reasoning benefit of graph-structured knowledge far more effectively than AR counterparts ($+0.058$ vs.\ $+0.026$ to $+0.038$ gain in Empathy \& Insight). Human evaluation with lay participants and licensed clinical psychologists, multi-judge automatic evaluation, and cross-layer probing analysis all converge to support these findings.

\subsection{Limitations and future work}

Despite strong performance, our approach has several limitations:
(1) diffusion-based generation incurs higher inference latency than autoregressive decoding (approximately 128 denoising steps vs.\ single-pass AR generation), which may limit real-time deployment;
(2) the BiGraph-RAG index is static, while real-world clinical knowledge evolves continuously---incremental graph update mechanisms are needed for long-term deployment;
(3) although we validated evaluation robustness with two judge models (Gemini-2.5-flash and GPT-4.1 Mini, $r = 0.981$) and confirmed that scores are stable with and without thought traces (Section~\ref{sec:synergy}), expanding to a broader multi-judge consensus framework would further strengthen evaluation reliability~\cite{badawi2026trust};
(4) the Chinese evaluation dataset CPsyCounE contains only 45 test dialogues across 9 topics---larger-scale, multi-lingual benchmarks are needed for comprehensive cross-cultural validation;
(5) our counseling-specific adaptation relies on LoRA fine-tuning with a fixed base model; fully pre-training a diffusion language model on domain-specific counseling data may unlock further performance.

Future work will explore model distillation to reduce inference latency, incremental graph update mechanisms for dynamic knowledge integration, multi-judge consensus frameworks with diverse LLM evaluators, and the construction of larger-scale multi-lingual counseling benchmarks. Extending BiGraph-Diffuse to more languages and cultural contexts, and investigating its applicability to adjacent domains such as legal consultation and educational tutoring, are promising directions.

\section*{Ethics statement}

This work develops a large language model for psychological counseling dialogue generation. We acknowledge the sensitive nature of mental health applications and have taken several precautions.

\paragraph{Data privacy and consent.} The OpenR1-Psy~\cite{hu2025beyond} dataset is constructed from publicly available, anonymized Reddit posts and existing psychological counseling datasets that have undergone rule-based filtering and manual review to remove personally identifiable information. CPsyCoun~\cite{zhang2024cpsycoun} is constructed from anonymized psychological counseling reports. The original authors implemented sanitization procedures including rule-based cleaning, manual rewriting, and human proofreading to remove personally identifiable information. We use CPsyCounD and CPsyCounE under their original license for non-commercial academic research. Both datasets are publicly available at \url{https://github.com/CAS-SIAT-XinHai/CPsyCoun} and \url{https://huggingface.co/datasets/CAS-SIAT-XinHai/CPsyCounR}.

\paragraph{Human subjects.} All 30 lay participants and three licensed clinical psychologists who took part in the human evaluation provided informed consent prior to participation.

\paragraph{Not a substitute for professional care.} Our model is designed as a research prototype and is not intended
to replace licensed mental health professionals. It does not perform clinical diagnosis or crisis intervention.

\paragraph{Safety and harm prevention.} We conducted multi-dimensional filtering during dataset construction to exclude harmful content. Evaluation metrics include a dedicated safety dimension, and both automatic and human evaluations confirm high safety scores.

\paragraph{Environmental impact.} Our fine-tuning uses LoRA adapters, requiring only a single 80\,GB GPU for approximately 9 hours. This is substantially more efficient than pre-training a model of comparable scale from scratch.

\paragraph{Reproducibility.} We will open-source our full pipeline to facilitate future research. Our code and model weights are publicly available at \url{https://github.com/Chekhov0919/BiGraph-Diffuse}.

\section*{Acknowledgments}
This work was supported by Jiangsu Province Frontier Program Project (BF2025036). They are also with Jiangsu Key Lab of Language Computing, Suzhou.

\bibliographystyle{IEEEtran}
\bibliography{reference}

@article{chung2023challenges,
  title={Challenges of large language models for mental health counseling},
  author={Chung, Neo Christopher and Dyer, George and Brocki, Lennart},
  journal={arXiv preprint arXiv:2311.13857},
  year={2023}
}

@inproceedings{badawi2026trust,
    title = "When Can We Trust {LLM}s in Mental Health? Large-Scale Benchmarks for Reliable {LLM} Evaluation",
    author = "Badawi, Abeer  and
      Rahimi, Elahe  and
      Laskar, Md Tahmid Rahman  and
      Grach, Sheri  and
      Bertrand, Lindsay  and
      Danok, Lames  and
      Dhanesh, Prathiba  and
      Huang, Jimmy  and
      Rudzicz, Frank  and
      Dolatabadi, Elham",
    editor = "Demberg, Vera  and
      Inui, Kentaro  and
      Marquez, Llu{\'i}s",
    booktitle = "Proceedings of the 19th Conference of the {E}uropean Chapter of the {A}ssociation for {C}omputational {L}inguistics (Volume 1: Long Papers)",
    month = mar,
    year = "2026",
    address = "Rabat, Morocco",
    publisher = "Association for Computational Linguistics",
    url = "https://aclanthology.org/2026.eacl-long.180/",
    doi = "10.18653/v1/2026.eacl-long.180",
    pages = "3873--3896",
    ISBN = "979-8-89176-380-7"
}

@misc{berglund2023reversal,
      title={The Reversal Curse: LLMs trained on "A is B" fail to learn "B is A"}, 
      author={Lukas Berglund and Meg Tong and Max Kaufmann and Mikita Balesni and Asa Cooper Stickland and Tomasz Korbak and Owain Evans},
      year={2024},
      eprint={2309.12288},
      archivePrefix={arXiv},
      primaryClass={cs.CL},
      url={https://arxiv.org/abs/2309.12288}, 
}

@article{bordin1979,
  title={The generalizability of the psychoanalytic concept of the working alliance.},
  author={Bordin, Edward S},
  journal={Psychotherapy: Theory, research \& practice},
  volume={16},
  number={3},
  pages={252},
  year={1979},
  publisher={Division of Psychotherapy (29), American Psychological Association}
}

@inproceedings{chen2023soulchat,
    title = "{S}oul{C}hat: Improving {LLM}s' Empathy, Listening, and Comfort Abilities through Fine-tuning with Multi-turn Empathy Conversations",
    author = "Chen, Yirong  and
      Xing, Xiaofen  and
      Lin, Jingkai  and
      Zheng, Huimin  and
      Wang, Zhenyu  and
      Liu, Qi  and
      Xu, Xiangmin",
    editor = "Bouamor, Houda  and
      Pino, Juan  and
      Bali, Kalika",
    booktitle = "Findings of the Association for Computational Linguistics: EMNLP 2023",
    month = dec,
    year = "2023",
    address = "Singapore",
    publisher = "Association for Computational Linguistics",
    url = "https://aclanthology.org/2023.findings-emnlp.83/",
    doi = "10.18653/v1/2023.findings-emnlp.83",
    pages = "1170--1183"
}

@article{deepseek2024deepseek,
  title={Deepseek-v3 technical report},
  author={Liu, Aixin and Feng, Bei and Xue, Bing and Wang, Bingxuan and Wu, Bochao and Lu, Chengda and Zhao, Chenggang and Deng, Chengqi and Zhang, Chenyu and Ruan, Chong and others},
  journal={arXiv preprint arXiv:2412.19437},
  year={2024}
}

@article{edge2024from,
  title={From local to global: A graph rag approach to query-focused summarization},
  author={Edge, Darren and Trinh, Ha and Cheng, Newman and Bradley, Joshua and Chao, Alex and Mody, Apurva and Truitt, Steven and Metropolitansky, Dasha and Ness, Robert Osazuwa and Larson, Jonathan},
  journal={arXiv preprint arXiv:2404.16130},
  year={2024}
}

@article{guo2024lightrag,
  title={Lightrag: Simple and fast retrieval-augmented generation},
  author={Guo, Zirui and Xia, Lianghao and Yu, Yanhua and Ao, Tian and Huang, Chao},
  journal={arXiv preprint arXiv:2410.05779},
  volume={2},
  number={3},
  year={2024}
}

@article{guo2024large,
  title={Large language models for mental health applications: systematic review},
  author={Guo, Zhijun and Lai, Alvina and Thygesen, Johan H and Farrington, Joseph and Keen, Thomas and Li, Kezhi},
  journal={JMIR mental health},
  volume={11},
  number={1},
  pages={e57400},
  year={2024},
  publisher={JMIR Publications Inc., Toronto, Canada}
}

@article{gutierrez2024hipporag,
  title={Hipporag: Neurobiologically inspired long-term memory for large language models},
  author={Guti{\'e}rrez, Bernal J and Shu, Yiheng and Gu, Yu and Yasunaga, Michihiro and Su, Yu},
  journal={Advances in neural information processing systems},
  volume={37},
  pages={59532--59569},
  year={2024}
}

@article{hu2021lora,
  title={Lora: Low-rank adaptation of large language models.},
  author={Hu, Edward J and Shen, Yelong and Wallis, Phillip and Allen-Zhu, Zeyuan and Li, Yuanzhi and Wang, Shean and Wang, Liang and Chen, Weizhu and others},
  journal={Iclr},
  volume={1},
  number={2},
  pages={3},
  year={2022}
}

@article{hu2025beyond,
  title={Beyond empathy: Integrating diagnostic and therapeutic reasoning with large language models for mental health counseling},
  author={Hu, He and Zhou, Yucheng and Si, Juzheng and Wang, Qianning and Zhang, Hengheng and Ren, Fuji and Ma, Fei and Cui, Laizhong and Tian, Qi},
  journal={arXiv preprint arXiv:2505.15715},
  year={2025}
}

@article{hurst2024gpt,
  title={Gpt-4o system card},
  author={Hurst, Aaron and Lerer, Adam and Goucher, Adam P and Perelman, Adam and Ramesh, Aditya and Clark, Aidan and Ostrow, AJ and Welihinda, Akila and Hayes, Alan and Radford, Alec and others},
  journal={arXiv preprint arXiv:2410.21276},
  year={2024}
}

@inproceedings{jiao2026prunerag,
  title={PruneRAG: Confidence-Guided Query Decomposition Trees for Efficient Retrieval-Augmented Generation},
  author={Jiao, Shuguang and Xiao, Xinyu and Wei, Yunfan and Qi, Shuhan and Huang, Chengkai and Sheng, Quan Z and Yao, Lina},
  booktitle={Proceedings of the ACM Web Conference 2026},
  pages={1923--1934},
  year={2026}
}

@article{liu2023chatcounselor,
  title={Chatcounselor: A large language models for mental health support},
  author={Liu, June M and Li, Donghao and Cao, He and Ren, Tianhe and Liao, Zeyi and Wu, Jiamin},
  journal={arXiv preprint arXiv:2309.15461},
  year={2023}
}

@inproceedings{lu2026mctsrzero,
  title={MCTSr-Zero: Self-Reflective Psychological Counseling Dialogues Generation via Principles and Adaptive Exploration},
  author={Lu, Hao and Gu, Yanchi and Huang, Haoyuan and Zhou, Yulin and Zhu, Ningxin and Li, Chen},
  booktitle={Proceedings of the AAAI Conference on Artificial Intelligence},
  volume={40},
  pages={32320--32328},
  year={2026}
}

@book{martin2013,
  title={Therapeutic presence: A mindful approach to effective therapy.},
  author={Geller, Shari M and Greenberg, Leslie S},
  year={2012},
  publisher={American Psychological Association}
}

@inproceedings{nguyen2025large,
  title={Do large language models align with core mental health counseling competencies?},
  author={Nguyen, Viet Cuong and Taher, Mohammad and Hong, Dongwan and Possobom, Vinicius Konkolics and Gopalakrishnan, Vibha Thirunellayi and Raj, Ekta and Li, Zihang and Soled, Heather J and Birnbaum, Michael L and Kumar, Srijan and others},
  booktitle={Findings of the Association for Computational Linguistics: NAACL 2025},
  pages={7488--7511},
  year={2025}
}

@article{nie2025large,
  title={Large language diffusion models},
  author={Nie, Shen and Zhu, Fengqi and You, Zebin and Zhang, Xiaolu and Ou, Jingyang and Hu, Jun and Zhou, Jun and Lin, Yankai and Wen, Ji-Rong and Li, Chongxuan},
  journal={arXiv preprint arXiv:2502.09992},
  year={2025}
}

@inproceedings{qi-etal-2025-kokorochat,
  title={Kokorochat: A japanese psychological counseling dialogue dataset collected via role-playing by trained counselors},
  author={Qi, Zhiyang and Kaneko, Takumasa and Takamizo, Keiko and Ukiyo, Mariko and Inaba, Michimasa},
  booktitle={Proceedings of the 63rd Annual Meeting of the Association for Computational Linguistics (Volume 1: Long Papers)},
  pages={12424--12443},
  year={2025}
}

@inproceedings{qiu-lan-2025-psydial,
  title={Psydial: A large-scale long-term conversational dataset for mental health support},
  author={Qiu, Huachuan and Lan, Zhenzhong},
  booktitle={Proceedings of the 63rd Annual Meeting of the Association for Computational Linguistics (Volume 1: Long Papers)},
  pages={21624--21655},
  year={2025}
}

@inproceedings{qiu2024smile,
  title={Smile: Single-turn to multi-turn inclusive language expansion via chatgpt for mental health support},
  author={Qiu, Huachuan and He, Hongliang and Zhang, Shuai and Li, Anqi and Lan, Zhenzhong},
  booktitle={Findings of the Association for Computational Linguistics: EMNLP 2024},
  pages={615--636},
  year={2024}
}

@article{rogers1957,
  title={The necessary and sufficient conditions of therapeutic personality change.},
  author={Rogers, Carl R},
  journal={Journal of consulting psychology},
  volume={21},
  number={2},
  pages={95},
  year={1957},
  publisher={American Psychological Association}
}

@article{sahoo2024simple,
  title={Simple and effective masked diffusion language models},
  author={Sahoo, Subham S and Arriola, Marianne and Schiff, Yair and Gokaslan, Aaron and Marroquin, Edgar and Chiu, Justin T and Rush, Alexander and Kuleshov, Volodymyr},
  journal={Advances in Neural Information Processing Systems},
  volume={37},
  pages={130136--130184},
  year={2024}
}

@article{shi2024simplified,
  title={Simplified and generalized masked diffusion for discrete data},
  author={Shi, Jiaxin and Han, Kehang and Wang, Zhe and Doucet, Arnaud and Titsias, Michalis},
  journal={Advances in neural information processing systems},
  volume={37},
  pages={103131--103167},
  year={2024}
}

@inproceedings{sun2021psyqa,
  title={Psyqa: A chinese dataset for generating long counseling text for mental health support},
  author={Sun, Hao and Lin, Zhenru and Zheng, Chujie and Liu, Siyang and Huang, Minlie},
  booktitle={Findings of the association for computational linguistics: ACL-IJCNLP 2021},
  pages={1489--1503},
  year={2021}
}

@inproceedings{xie2025psydt,
  title={Psydt: Using llms to construct the digital twin of psychological counselor with personalized counseling style for psychological counseling},
  author={Xie, Haojie and Chen, Yirong and Xing, Xiaofen and Lin, Jingkai and Xu, Xiangmin},
  booktitle={Proceedings of the 63rd Annual Meeting of the Association for Computational Linguistics (Volume 1: Long Papers)},
  pages={1081--1115},
  year={2025}
}

@inproceedings{yang2024mentallama,
  title={MentaLLaMA: interpretable mental health analysis on social media with large language models},
  author={Yang, Kailai and Zhang, Tianlin and Kuang, Ziyan and Xie, Qianqian and Huang, Jimin and Ananiadou, Sophia},
  booktitle={Proceedings of the ACM Web Conference 2024},
  pages={4489--4500},
  year={2024}
}

@inproceedings{yang2025cami,
  title={Cami: A counselor agent supporting motivational interviewing through state inference and topic exploration},
  author={Yang, Yizhe and Achananuparp, Palakorn and Huang, He-Yan and Jiang, Jing and Kit, Phey Ling and Lim, Nicholas Gabriel and Ern, Cameron Tan Shi and Lim, Ee-Peng},
  booktitle={Proceedings of the 63rd Annual Meeting of the Association for Computational Linguistics (Volume 1: Long Papers)},
  pages={21037--21081},
  year={2025}
}

@inproceedings{Yin_Li_Zhang_Wang_Shao_Li_Chen_Jiang_2025,
  title={Mdd-5k: A new diagnostic conversation dataset for mental disorders synthesized via neuro-symbolic llm agents},
  author={Yin, Congchi and Li, Feng and Zhang, Shu and Wang, Zike and Shao, Jun and Li, Piji and Chen, Jianhua and Jiang, Xun},
  booktitle={Proceedings of the AAAI Conference on Artificial Intelligence},
  volume={39},
  pages={25715--25723},
  year={2025}
}

@inproceedings{zhang2024cpsycoun,
  title={Cpsycoun: A report-based multi-turn dialogue reconstruction and evaluation framework for chinese psychological counseling},
  author={Zhang, Chenhao and Li, Renhao and Tan, Minghuan and Yang, Min and Zhu, Jingwei and Yang, Di and Zhao, Jiahao and Ye, Guancheng and Li, Chengming and Hu, Xiping},
  booktitle={Findings of the Association for Computational Linguistics: ACL 2024},
  pages={13947--13966},
  year={2024}
}

@inproceedings{zhang2024escot,
  title={Escot: Towards interpretable emotional support dialogue systems},
  author={Zhang, Tenggan and Zhang, Xinjie and Zhao, Jinming and Zhou, Li and Jin, Qin},
  booktitle={Proceedings of the 62nd Annual Meeting of the Association for Computational Linguistics (Volume 1: Long Papers)},
  pages={13395--13412},
  year={2024}
}

@inproceedings{zhu2026psiarena,
  title={$\Psi$-arena: Interactive assessment and optimization of llm-based psychological counselors with tripartite feedback},
  author={Zhu, Shijing and Chen, Zhuang and Bi, Guanqun and Li, Binghang and Deng, Yaxi and Wan, Dazhen and Peng, Libiao and Xiao, Xiyao and Zhang, Rongsheng and Lv, Tangjie and others},
  booktitle={Proceedings of the AAAI Conference on Artificial Intelligence},
  volume={40},
  pages={2272--2280},
  year={2026}
}

@article{zhuang2025linearrg,
  title={Linearrag: Linear graph retrieval augmented generation on large-scale corpora},
  author={Zhuang, Luyao and Chen, Shengyuan and Xiao, Yilin and Zhou, Huachi and Zhang, Yujing and Chen, Hao and Zhang, Qinggang and Huang, Xiao},
  journal={arXiv preprint arXiv:2510.10114},
  year={2025}
}

@article{zhu2026uncertainty,
  title={Uncertainty-aware multimodal affective data fusion for personalized mental health dialogue},
  author={Zhu, Xianxun and Cambria, Erik and Chen, Hui},
  journal={Pattern Recognition},
  pages={114574},
  year={2026},
  publisher={Elsevier}
}

@article{wulanguage,
  title={Language-Centered Mental Healthcare with Large Language Models: A Comprehensive Survey Across the Care Continuum},
  author={Wu, Jialun and He, Kai and Kampman, Onno P and Gerard, Chung Siew Keong and Wilson, Goh Wen Bin and Lin, Qika and Xu, Jiaxing and Du, Yanrui and Shang, Xuequn and Cambria, Erik and others},
  journal={SSRN},
  year={2026}
}

@article{ji2023rethinking,
  title={Rethinking large language models in mental health applications},
  author={Ji, Shaoxiong and Zhang, Tianlin and Yang, Kailai and Ananiadou, Sophia and Cambria, Erik},
  journal={arXiv preprint arXiv:2311.11267},
  year={2023}
}

@article{ji2023domain,
  title={Domain-specific continued pretraining of language models for capturing long context in mental health},
  author={Ji, Shaoxiong and Zhang, Tianlin and Yang, Kailai and Ananiadou, Sophia and Cambria, Erik and Tiedemann, J{\"o}rg},
  journal={arXiv preprint arXiv:2304.10447},
  year={2023}
}

\end{document}